\documentclass[runningheads]{llncs}
\usepackage[mobile]{eccv}

\usepackage{eccv}

\usepackage[mobile]{eccv}

\usepackage[utf8]{inputenc}
\usepackage[T1]{fontenc}
\usepackage{eccvabbrv}
\usepackage{graphicx}
\usepackage{booktabs}

\usepackage[accsupp]{axessibility}

\usepackage[pagebackref,breaklinks,colorlinks,citecolor=eccvblue]{hyperref}

\usepackage{float}

\usepackage{orcidlink}
\usepackage{colortbl}

\usepackage{xcolor}
\usepackage{arydshln}
\usepackage{multirow}
\definecolor{mygray}{gray}{.92}
\usepackage{pifont}

\newcommand{\figref}[1]{Fig. \ref{#1}}

\usepackage{algorithm}
\usepackage{algpseudocode}

\newcommand{\modelname}{FGOS-Net}
\usepackage[misc]{ifsym}
\begin{document}

\title{Bridging the Geometry Mismatch: Frequency-Aware Anisotropic Serialization for Thin-Structure SSMs}

\renewcommand\footnotemark{}
\author{Jin Bai $^{1,2}$  \and
Huiyao Zhang $^{1,2}$\and  
Qi Wen $^{2}$ $^{(\textrm{\Letter})}$ \and 
Ningyang Li $^{1,2}{1,2}$ \and 
Shengyang Li $^2$  \and 
Atta ur Rahman$^{3}$ \and 
Xiaolin Tian$^{4}$
   \\
\thanks{$\textrm{\Letter}$~Corresponding authors.}
}

\authorrunning{}

\institute{ $^1$University of Chinese Academy of Sciences, \\ $^2$Technology and Engineering Center for Space Utilization, Chinese Academy of Sciences \\$^3$ University of Peshawar, Pakistan\\$^4$ Macau University of Science and Technology, China\\
\email{baijin25@mails.ucas.ac.cn,zhanghuiyao25@csu.ac.cn, wenqi@csu.ac.cn}}

\maketitle
% =========================
% Abstract
% =========================
\begin{abstract}
The segmentation of thin linear structures is inherently topology\allowbreak-critical, where minor local errors can sever long-range connectivity. While recent State-Space Models (SSMs) offer efficient long-range modeling, their isotropic serialization (e.g., raster scanning) creates a geometry mismatch for anisotropic targets, causing state propagation across rather than along the structure trajectories. To address this, we propose FGOS-Net, a framework based on frequency\allowbreak-geometric disentanglement. We first decompose features into a stable topology carrier and directional high-frequency bands, leveraging the latter to explicitly correct spatial misalignments induced by downsampling. Building on this calibrated topology, we introduce frequency-aligned scanning
that elevates serialization to a geometry-conditioned decision, preserving
direction-consistent traces. Coupled with an active probing strategy to
selectively inject high-frequency details and suppress texture ambiguity,
FGOS-Net consistently outperforms strong baselines across four challenging
benchmarks. Notably, it achieves 91.3\% mIoU and 97.1\% clDice on DeepCrack
while running at 80 FPS with only 7.87 GFLOPs.

\end{abstract}

% =========================
% Introduction
% =========================

\section{Introduction}
\label{sec:intro}

Thin-structure segmentation presents a persistent challenge in computer vision due to the extreme anisotropy and topological sensitivity of targets such as cracks, vessels, and roads~\cite{deepcrack,crackmap,crack500,scsegamba}.

Among these, cracks represent a particularly challenging case due to highly variable widths, severe clutter, and low signal-to-noise ratios, which significantly hinder long-range connectivity preservation. While this mismatch affects thin structures broad-\allowbreak{}ly, we focus on crack segmentation as an extreme testbed to rigorously isolate serialization failures from confounding factors.

State-Space Models (SSMs) have emerged as an efficient paradigm for long-range dependency modeling~\cite{s4,mamba}.
However, most SSM-based segmentation frameworks rely on isotropic serialization (e.g., raster scanning) to flatten 2D feature maps~\cite{vmamba_neurips}.
This assumption breaks for anisotropic linear geometry, creating a serialization-induced geometry mismatch: the recurrent state is forced to traverse orthogonally to the structure's dominant axis, fragmenting what should be a single continuous trace into disjoint segments.
While recent lightweight SSMs like SCSegamba~\cite{scsegamba} minimize parameters, their complex scan patterns limit throughput. We instead prioritize FLOPs and FPS as the true deployment bottleneck.

To bridge this gap, we propose \modelname{}, a lightweight framework that aligns 1D state-space modeling with 2D anisotropic geometry through Frequency-Geometric Disentanglement.
Leveraging the Haar Discrete Wavelet Transform (DWT) as a geometric prior, we decompose features into a stable low-frequency structural support and directional high-frequency bands.
Building on this decomposition, we introduce Frequency-Aligned Scanning (FA-Scan), which elevates serialization from a fixed operation to a geometry-conditioned modeling decision by assigning sub-band-aligned traversal trajectories.
Complementarily, to suppress texture-induced ambiguity, we propose Active Spectral-Geometric Probing (ASGP).
This module actively evolves probes on the topology carrier to generate a topology-conditioned gate, selectively injecting high-frequency details only when validated by structural consistency.
These are integrated into a hierarchical encoder with efficient LightGate Bottleneck (LGB) gating and a parallel GFA decoder for multi-scale fusion.

Our contributions are summarized as follows:
\begin{itemize}
    \item We identify isotropic serialization as an architectural blind spot in applying State-Space Models to thin-structure segmentation, where fixed scan orders disrupt structure-consistent long-range propagation.
    \item We propose \modelname{}, a frequency--geometric disentanglement framework that explicitly replaces passive frequency fusion with geometry-aligned serialization and topology-conditioned detail injection via FA-Scan and ASGP, enabling robust boundary modeling under cluttered backgrounds.
    \item Extensive experiments on four benchmarks demonstrate that \modelname{} consistently improves boundary integrity and connectivity preservation over recent CNN-, SSM-, and topology-specialized baselines~\cite{rind,ctcrack,simcrack,ffm,glcp,vmamba_neurips,swinumamba,umamba,plainmamba}, while maintaining a highly competitive accuracy--efficiency trade-off.
\end{itemize}

% =========================
\section{Related Work}
\label{sec:rw}

\noindent\textbf{Anisotropic Linear Structure Segmentation.}
Extracting thin linear structures (\eg cracks, vessels, roads) is inherently topology-dominated: even minor pixel errors can sever long-range connectivity~\cite{mosinska2018,cldice}.
Existing topological losses or skeleton-based constraints~\cite{cldice,glcp} serve as supervisory regularizers but do not alter the feature modeling itself.
Recent self-similarity approaches~\cite{ffm} leverage fractal statistics to capture structural recurrence, yet remain agnostic to serialization order.
Architecturally, most CNN- and Transformer-based models operate with isotropic kernels or global attention, leaving them vulnerable to (i) connectivity discontinuity and (ii) texture leakage~\cite{strip_pooling,csanet,cs2net,topology_mamba,mambavesselplus,serpmamba}.

\noindent\textbf{State-Space Models and Serialization Mismatch.}
Visual SSMs achieve linear-complexity long-range modeling via selective scan~\cite{mamba,mamba2,vmamba_neurips,vim,localmamba,plainmamba,spatialmamba,mila,vssd}, yet their serialization of 2D maps into 1D sequences is typically data-agnostic~\cite{vmamba_neurips,swinumamba,umamba,mamba_nd,defmamba,groupmamba,damamba}.
For thin structures this creates a fundamental geometry mismatch: spatially adjacent points along a structure may map to distant 1D positions when the scan cuts orthogonally across the dominant axis, fracturing continuous traces.
We argue that the traversal order should be informed by the underlying geometry, utilizing directional priors from frequency decomposition.

\noindent\textbf{Frequency-Aware Learning and Gated Injection.}
Frequency decomposition (\eg, Wavelet or Fourier) disentangles low-frequency semantics from high-frequency boundary cues~\cite{dwt_seg}, yet high-frequency bands encode both meaningful edges and spurious textures.
Most wavelet-based networks adopt passive fusion (\eg, concatenation), indiscriminately injecting all high-frequency content~\cite{gatedscnn,gff,fcanet}.
Our approach instead redefines the low-frequency component as a stable topology carrier that actively regulates detail injection via global topological agreement.

\noindent\textbf{Efficient Architecture Design.}
Lightweight designs~\cite{mobilenet} sacrifice high-frequen-cy detail through aggressive downsampling, particularly harmful for thin structures~\cite{unet,resnet}.
Our framework combines stage-adaptive LightGate bottleneck (LGB) with parallel multi-scale fusion (GFA), preserving boundary detail without additional latency.

\section{Method}
\label{sec:method}

\subsection{Overall Architecture}
\label{subsec:overall}

\begin{figure*}[t]
  \centering
  \includegraphics[width=\textwidth]{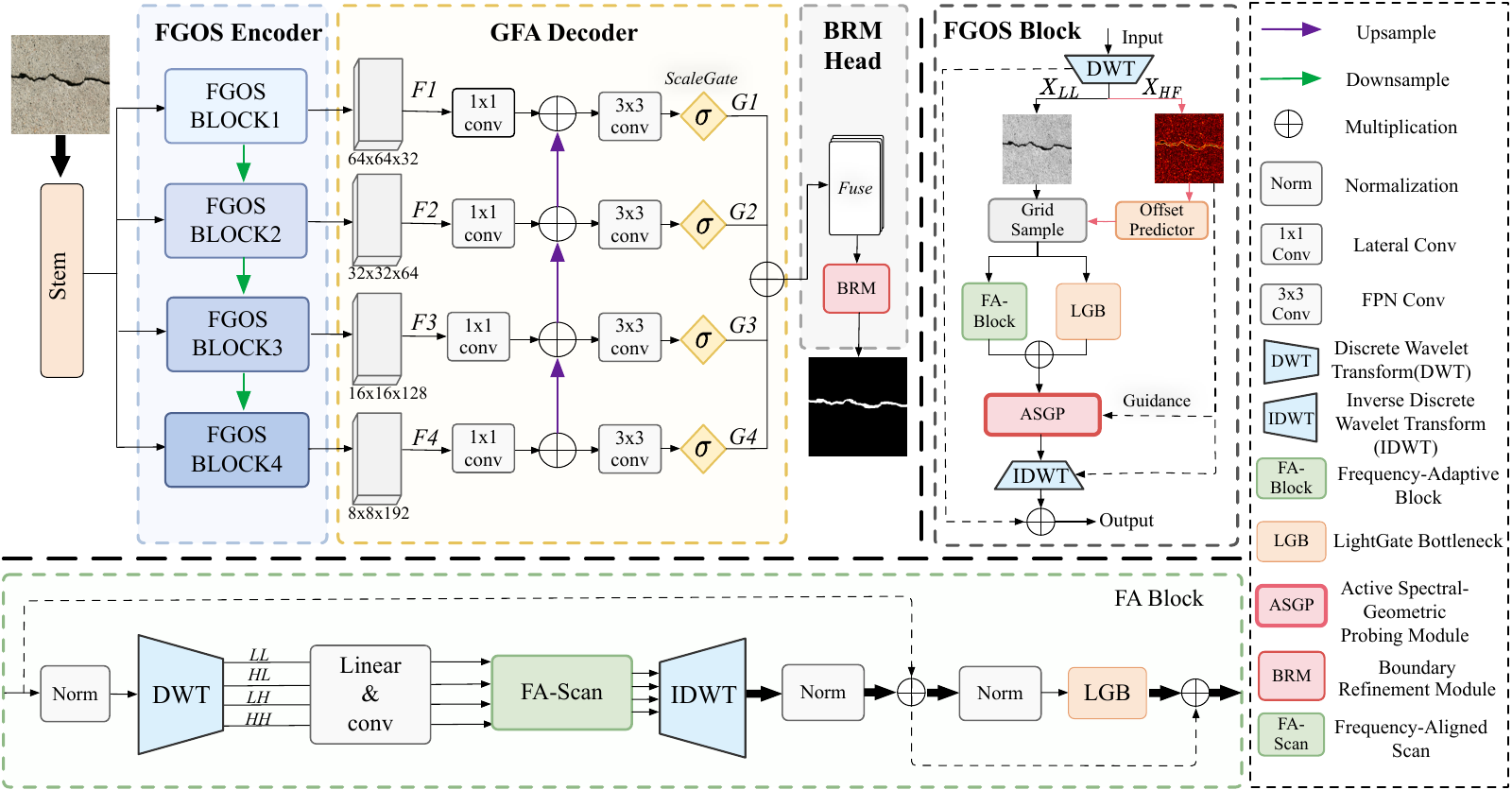}
\caption{\textbf{Overview of \modelname{}.}
Each FGOS block performs: (1) DWT-based frequency disentanglement into topology carrier $X_{LL}$ and directional details $X_{HF}$;
(2) detail-driven alignment of $X_{LL}$;
(3) geometry-aligned FA-Scan for topology modeling and ASGP for topology-conditioned detail gating;
(4) IDWT-based reconstruction.
A parallel GFA decoder and BRM head fuse multi-scale features and refine boundaries.}
  \label{fig:main_arch}
\end{figure*}

\noindent\textbf{Problem Formulation.}
Standard SSMs serialize 2D features via isotropic strategies (\eg, raster),
implicitly assuming spatial adjacency is preserved under a fixed 1D
traversal~\cite{segman,spatialmamba}. For anisotropic thin structures, this
breaks down: adjacent points along the structure may map to distant 1D positions.

\noindent\textbf{Core Principle: Split, Align, and Model.}
To correct this modeling-level geometry mismatch, \modelname{} adopts a frequency--geometric disentanglement paradigm that decouples topology modeling from noisy boundary details. Each FGOS stage (\cref{fig:main_arch}, right) processes features $X$ through four steps:

\noindent\textbf{Disentangle (DWT).} We employ an invertible Haar DWT to physically separate the input into a low-frequency topology carrier ($X_{LL}$) and directional high-frequency bands:
\begin{equation}
    \{X_{LL}, X_{LH}, X_{HL}, X_{HH}\} = \mathrm{DWT}(X), \quad X_{HF} = \{X_{LH}, X_{HL}, X_{HH}\}.
\end{equation}
We adopt the Haar wavelet for three reasons:
(i)~its $2{\times}2$ support yields the sharpest directional decomposition
($LH$/$HL$ encode pure horizontal/vertical gradients) with minimal
cross-structure contamination, unlike smoother wavelets (\eg, Daubechies)
whose wider support mixes adjacent structures;
(ii)~perfect invertibility guarantees zero information loss upon IDWT
reconstruction;
(iii)~its minimal arithmetic cost is critical for real-time throughput.
Importantly, DWT operates on multi-channel feature maps rather than raw
pixels: sub-pixel structural cues have already been distributed across
channels by the convolutional stem, and the subsequent IDWT restores
full spatial resolution within each stage.

\noindent\textbf{Align (Geometric Prior).} Crucially, we avoid directly fusing $X_{HF}$ into the topology stream. Instead, we aggregate the directional bands to predict a light-\allowbreak{}weight deformation field $\Delta \mathbf{p}$, which spatially aligns the topology carrier $X_{LL}$ to the standard coordinate grid $\mathbf{p}$ via \texttt{GridSample}:
\begin{equation}
    \tilde{X}_{LL} = \mathrm{GridSample}\left(X_{LL}, \, \mathbf{p} + \mathcal{P}(\mathrm{Concat}[X_{HF}])\right),
\end{equation}
where $\mathcal{P}$ denotes a lightweight two-layer $3{\times}3$ convolutional projection predicting a 2-channel offset field. Progressive downsampling introduces sub-pixel quantization errors that shift the topology carrier from the true
structural locus.
The directional HF bands retain positional cues about where structures
reside, enabling the predicted deformation field to correct spatial
misregistration of $X_{LL}$ without injecting texture noise---analogous to
edge-guided sub-pixel alignment.

\noindent\textbf{Model \& Gate.} The streams are then processed in parallel:

\noindent\textbf{Topology Stream ($\tilde{X}_{LL}$):} Processed by the FA-Block (Sec.~\ref{subsec:fa_encoder}), which assigns sub-band-aligned serialization trajectories (\eg, Horizontal/Vertical/Hilbert) to preserve geometric continuity during SSM propagation.

\noindent\textbf{Detail Stream ($X_{HF}$):} Processed by ASGP (Sec.~\ref{subsec:asgp}), which actively evolves probes on $\tilde{X}_{LL}$ to generate a spatial gate $M$, ensuring that only structure-consistent high-frequency details are preserved.

\noindent\textbf{Merge (IDWT).} Finally, the refined topology and gated details are recombined via Inverse DWT to restore the full-spectrum feature map for the next stage.

%==================================================
\subsection{The FA-Block: Geometry-Aligned Modeling}
\label{subsec:fa_encoder}

\begin{figure}[t]
  \centering
  \includegraphics[width=1\textwidth]{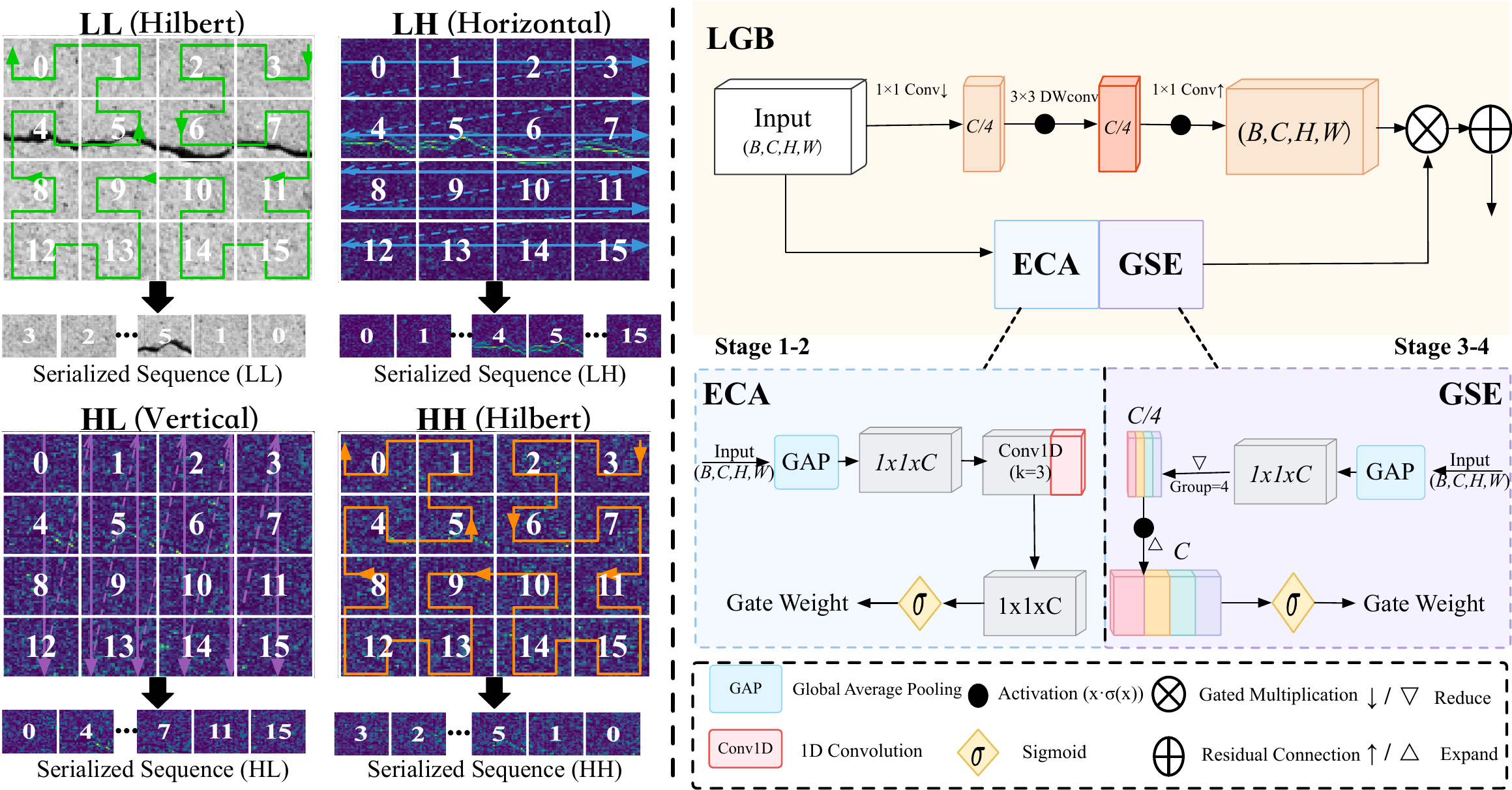}
\caption{\textbf{FA-Block components.}
Left: FA-Scan assigns deterministic, sub-band-aligned serialization trajectories (Horizontal/Vertical for directional bands, Hilbert for isotropic bands).
Right: The LightGate Bottleneck (LGB) employs a symmetric bottleneck ($C{\to}C/4{\to}C$) modulated by stage-adaptive gating (ECA~\cite{eca} / GSE~\cite{senet}), reducing parameters by $\approx 94\%$ vs. standard FFNs. ECA and GSE denote Efficient Channel Attention and Grouped Squeeze-and-Excitation, respectively.}
  \label{fig:fa_lgb}
\end{figure}

\noindent\textbf{FA-Scan: Sub-band-Aligned Serialization.}
To obtain geometry-aware trajectories without introducing global high-frequency noise, we perform a local Haar DWT exclusively on the aligned topology carrier $\tilde{X}_{LL}$, yielding four internal sub-bands: $\{ \tilde{X}_{LL}^{ll}, \tilde{X}_{LL}^{lh}, \tilde{X}_{LL}^{hl}, \tilde{X}_{LL}^{hh} \} = \mathrm{DWT}(\tilde{X}_{LL})$. This transforms the serialization from isotropic raster scanning into a sub-band conditioned task. Let $\tilde{X}_{LL}^{sub}$ denote these local features. We assign deterministic trajectories $\mathcal{S}_{sub}$ matching the dominant geometry of each band:
\begin{equation}
\mathcal{S}_{lh} = \mathcal{S}_{\text{Horz}}, \quad
\mathcal{S}_{hl} = \mathcal{S}_{\text{Vert}}, \quad
\mathcal{S}_{ll} = \mathcal{S}_{hh} = \mathcal{S}_{\text{Hilbert}}.
\label{eq:scan_logic}
\end{equation}
\vspace{-2pt}
\noindent Here, $\mathcal{S}_{\text{Hilbert}}$ denotes the Hilbert space-filling curve~\cite{hilbert}. Since IDWT reconstructs the signal by summing orthogonal components, structures at arbitrary orientations (\eg, $45^\circ$ cracks) are recovered through the complementary scanned sequences.

To maximize parameter efficiency, the serialized sequences are processed by a shared SSM operator $\Psi$ (\eg, VSS Block~\cite{vmamba_neurips}), differing only in traversal order:
\begin{equation}
\hat{X}_{LL}^{sub} = \mathcal{S}_{sub}^{-1}\!\left(\Psi\!\left(\mathcal{S}_{sub}(\tilde{X}_{LL}^{sub})\right)\right), \quad Y_{\text{scan}} = \mathrm{IDWT}\big(\{\hat{X}_{LL}^{sub}\}\big).
\label{eq:fascan_process}
\end{equation}
\textbf{Intuition:} $\mathcal{S}_{\text{Horz}}$ ensures horizontal structures in $\tilde{X}_{LL}^{lh}$ remain contiguous in 1D. For isotropic bands ($\tilde{X}_{LL}^{ll}, \tilde{X}_{LL}^{hh}$), the Hilbert curve minimizes linearization distance, preserving 2D locality better than raster scans.
Since $\mathcal{S}$ is a fixed indexing operation, FA-Scan adds no learnable parameters and uses precomputed Hilbert indexing.

\noindent\textbf{LGB: Efficient Selective Mixing.}
We replace heavy FFNs with the LightGate Bottleneck (LGB) (Fig.~\ref{fig:fa_lgb}, Right).
It comprises a symmetric bottleneck path $\mathcal{F}(Y)$ ($C{\to}C/4{\to}C$) and a stage-adaptive gate $\mathcal{G}_s(Y)$.
The output is defined as $\mathrm{LGB}(Y) = Y + \mathcal{G}_s(Y) \odot \mathcal{F}(Y)$.
Crucially, the gating policy $\mathcal{G}_s$ adapts to the feature hierarchy:
\begin{equation}
\label{eq:lgb_gating}
\mathcal{G}_s(Y) = \sigma\!\left(
\begin{cases}
\mathrm{Conv1D}_k(\mathrm{GAP}(Y)), & s \in \{1, 2\} \quad (\text{ECA}), \\
\mathbf{W}_2(\delta(\mathbf{W}_1(\mathrm{GAP}(Y)))), & s \in \{3, 4\} \quad (\text{GSE}),
\end{cases}
\right)
\end{equation}
where $\mathcal{F}$ employs a $3{\times}3$ DWConv sandwiched between projections ($r{=}4$). This hybrid design reduces parameters by $\approx 94\%$ while retaining representational capacity.

\subsection{Active Spectral-Geometric Probing}
\label{subsec:asgp}

\begin{figure}[t]
  \centering
  \includegraphics[width=0.80\textwidth]{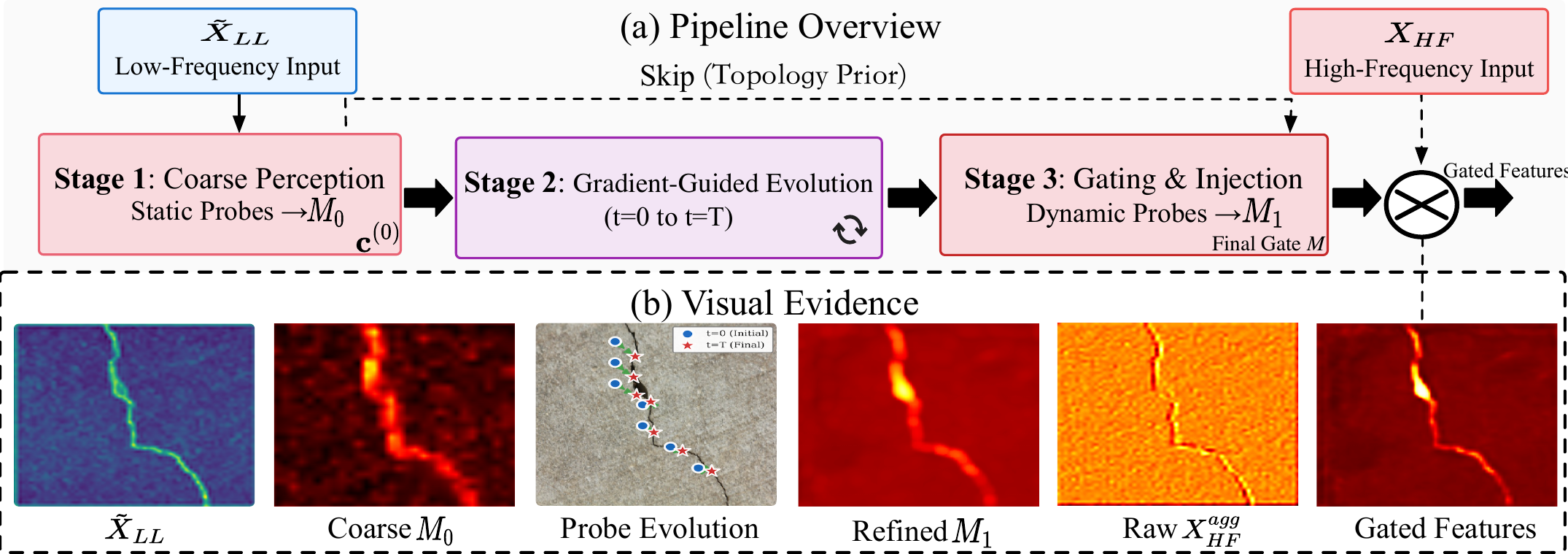} 

\caption{\textbf{Active Spectral-Geometric Probing (ASGP).}
The module resolves high-frequency ambiguity via:
(i) Coarse Perception: Initializing probes on the topology carrier $\tilde{X}_{LL}$;
(ii) Gradient-Guided Evolution: Iteratively refining probe positions ($t=0{\to}T$) via differentiable gradient ascent on the potential field $M_0$;
(iii) Gating \& Injection: Generating a topology-validated mask $M$ to selectively gate high-frequency details ($X_{HF} \odot M$).}
  \label{fig:asgp}

\end{figure}

\begin{algorithm}[t]
\caption{Gradient-Guided Probe Evolution}
\label{alg:asgp}
\small
\textbf{Input:} Coarse map $M_0$, Feature $\tilde{X}_{LL}$, Init coords $\mathcal{C}^{(0)}$, Steps $T$ \\
\textbf{Hyperparams:} Radius $\delta=0.15$, $\lambda_1{=}0.1$, $\lambda_2{=}0.05$, $\epsilon=1e^{-5}$
\begin{algorithmic}[1]
\For{$t = 0$ \textbf{to} $T-1$}
    \State $\mathbf{f}_i^{(t)} \gets \text{GridSample}(\tilde{X}_{LL}, \mathcal{C}^{(t)})$ \Comment{Feature Extraction}
    \State $\Delta\mathbf{c}_{\text{sem}} \gets \text{MLP}(\mathbf{f}_i^{(t)})$ \Comment{Semantic Offset}
    \State $\mathbf{g}_i \gets \nabla_{\mathcal{C}}\,\text{GridSample}(M_0, \mathcal{C}^{(t)})$ \Comment{Differentiable Interpolation}
    \State Compute pairwise distances $d_{ij}=\lVert \mathbf{c}_i-\mathbf{c}_j\rVert_2$
    \State $\mathcal{F}_{div}^{(t)} \gets \sum_{j\neq i} \frac{\mathbf{c}_i-\mathbf{c}_j}{d_{ij}+\epsilon}\cdot \text{ReLU}\!\left(1-\frac{d_{ij}}{\delta}\right)$ \Comment{Diversity Constraint}
    \State $\mathcal{C}^{(t+1)} \gets \text{Clamp}\!\left( \mathcal{C}^{(t)}+\Delta\mathbf{c}_{\text{sem}}+\lambda_1\mathbf{g}_i+\lambda_2\mathcal{F}_{div}^{(t)}, -1,1\right)$
\EndFor
\State \textbf{Return} refined coordinates $\mathcal{C}^{(T)}$
\end{algorithmic}
\end{algorithm}

\noindent\textbf{Motivation: Topology-Conditioned Gating.}
Wavelet high-frequency bands ($X_{HF}$) contain both structural boundaries and background texture gradients. Naively injecting them into the semantic stream corrupts topology reasoning with noise. ASGP addresses this by formulating detail injection as a conditional gating problem: we actively localize structure-supporting regions on the stable topology carrier $\tilde{X}_{LL}$ to filter $X_{HF}$. Unlike passive attention, we employ an iterative optimization process to validate high-frequency evidence against global connectivity. Unlike deformable attention~\cite{dcn,deformable_detr}, which learns offsets purely
from data, ASGP guides probe migration via an explicit geometric potential
field ($M_0$) and enforces spatial diversity through a physics-inspired
repulsion constraint---properties absent in standard deformable mechanisms.

\noindent\textbf{Stage 1: Coarse Perception.}
We initialize $N$ static learnable probes $\{\mathbf{c}_i^{(0)}\}$ to query $\tilde{X}_{LL}$ with linear complexity $\mathcal{O}(NHW)$. These probes generate attention maps aggregated into a coarse potential field $M_0 = \sigma(\frac{1}{N}\sum_{i=1}^{N}\mathrm{Attn}(\mathbf{q}_i, \tilde{X}_{LL}))$. High responses in $M_0$ indicate probable structural regions, serving as the initialization landscape for probe refinement.
With $N{=}64 \ll H{\times}W$, the iterative evolution incurs negligible
overhead versus dense attention.

\noindent\textbf{Stage 2: Gradient-Guided Probe Evolution.}
We perform a sparse-to-dense evolution over $T$ steps. The update rule combines semantic offsets with gradient guidance derived from the first-order topology map $M_0$ (via bilinear interpolation):
\begin{equation}
\label{eq:evolution}
\mathbf{c}_i^{(t+1)} = \mathcal{P}_{\Omega}\Big(
\mathbf{c}_i^{(t)}
+ \Delta\mathbf{c}_{sem}^{(t)}
+ \lambda_1 \nabla_{\mathbf{c}} M_0(\mathbf{c}_i^{(t)})
+ \lambda_2 \mathcal{F}_{div, i}^{(t)}
\Big).
\end{equation}
Here, $\mathcal{P}_{\Omega}$ projects coordinates onto $[-1, 1]$. The update is driven by three terms: 
\textbf{1) Semantic Offset.} A learnable shift $\Delta\mathbf{c}_{sem}$ predicted via MLP from local features $\mathbf{f}_i^{(t)}$, allowing data-driven adjustments. 
\textbf{2) Gradient Ascent.} We treat $M_0$ as a continuous potential field. The term $\nabla_{\mathbf{c}} M_0(\mathbf{c}_i)$ drives probes to "climb" the probability peaks of the topological structure. 
\textbf{3) Diversity Constraint.}
Pure gradient ascent inherently drives probes towards a few global maxima (e.g., the widest crack segments), leading to spatial degeneracy.
To counteract this, we introduce a truncated repulsion force
$\mathcal{F}_{div} = \sum_{j \neq i} \frac{\mathbf{c}_i - \mathbf{c}_j}{\|\mathbf{c}_i - \mathbf{c}_j\|_2 + \epsilon} \cdot \max(0, 1 - \frac{\|\mathbf{c}_i - \mathbf{c}_j\|_2}{\delta})$.
This truncated repulsion enforces a minimum separation $\delta$, dispersing
probes along curvilinear ridges of $M_0$ rather than clustering at
high-contrast centroids.
Since all probe coordinates operate in the normalized $[-1, 1]$ space,
$\delta{=}0.15$ represents a fixed fraction of the feature map
extent and is inherently resolution-invariant, requiring no manual adjustment
across input scales or structure widths.

\noindent\textbf{Stage 3: Gating and Injection.}
After $T$ steps, the refined probe states are projected back to pixel space to form the fine-grained mask $M_1$. The final spatial gate $M = \sigma(w M_1 + (1-w) M_0)$ (with $w{=}0.5$) modulates the high-frequency input via $X_{HF}^{agg} = X_{HF} \odot M$, injecting detail evidence only where topology agrees.

\subsection{Reconstruction and Refinement}
\label{subsec:reconstruction}

\noindent\textbf{Motivation: semantic--spatial mismatch across pyramid levels.}
Deep features are semantically strong but spatially coarse, while shallow features are spatially precise yet noisy.
A sequential top-down decoder can accumulate smoothing errors.
We instead use a parallel multi-scale fusion with explicit scale calibration, followed by a lightweight boundary refiner.

\noindent\textbf{Gated Feature Aggregation (GFA).}
Given encoder features $\{F_i\}_{i=1}^{4}$, we project them to a shared embedding and upsample to the spatial size of $F_1$:
\begin{equation}
\tilde{F}_i = \mathcal{U}_{bilinear}\big(\phi_i(F_i),\, \text{size}(F_1)\big).
\label{eq:gfa_proj}
\end{equation}
Unlike sequential FPN-style fusion, which accumulates smoothing errors across
levels, we calibrate each scale independently via a channel-wise ScaleGate.
Let $\mathbf{W}_1, \mathbf{W}_2$ be channel-reduction layers.
The gating vector $s_i \in \mathbb{R}^{1 \times C \times 1 \times 1}$ is computed as:
\begin{equation}
s_i=\sigma\!\left(\mathbf{W}_2\cdot \mathrm{ReLU}\!\left(\mathbf{W}_1\cdot \mathrm{AvgPool}(\tilde{F}_i)\right)\right),\quad
G_i=s_i\odot \tilde{F}_i,
\label{eq:scalegate}
\end{equation}
and fuse in one shot:
\begin{equation}
F_{\text{fused}}=\mathrm{Conv}_{3\times3}\left(\sum_{i=1}^{4}G_i\right).
\label{eq:gfa_fuse}
\end{equation}

\noindent\textbf{Boundary Refinement Module (BRM).}
To counteract boundary smoothing from upsampling, BRM adopts a dual-branch residual design:
a context branch (depthwise-separable convolutions) improves semantic consistency, and an edge branch (pure depthwise convolution) preserves high-frequency gradients.
The two outputs are concatenated, projected to the original channel dimension, and added back to $F_{\text{fused}}$ to sharpen boundaries without disrupting global topology.

\section{Experiments}
\label{sec:exp}

\subsection{Experimental Setup}
\label{subsec:setup}
\noindent\textbf{Datasets.} We evaluate on four benchmarks: 
DeepCrack~\cite{deepcrack} (537 images) and Crack500~\cite{crack500} (3,368 images) featuring low-contrast topology; 
CrackMap~\cite{crackmap} (120 images) dominated by texture noise (\eg, water or oil stains); 
and TUT~\cite{scsegamba} (1,408 images) spanning eight
materially diverse environments---bitumen, cement, bricks, plastic
runways, tiles, metal surfaces, generator blades, and underground
pipelines---that introduce significant cross-scenario variation in
texture, lighting, and crack morphology, effectively serving as a
multi-domain benchmark within the thin-structure category.

\noindent\textbf{Implementation \& Metrics.}
Models follow official splits and are trained in PyTorch on a single NVIDIA RTX 3090 for 100 epochs with AdamW ($lr{=}1{\times}10^{-4}$, cosine decay) and $\mathcal{L}_{BCE}$+$\mathcal{L}_{Dice}$.
Inputs are resized to $256{\times}256$ with random flip, rotate (${\pm}15^\circ$), and photometric distortion.
ASGP uses $T{=}3, N{=}64$.
We report mIoU, F1, Precision, Recall, boundary-aware ODS, and connectivity-aware clDice~\cite{cldice}. FPS is measured at $256{\times}256$ including ASGP iterations.

\begin{table*}[t]
\centering
\caption{\textbf{Quantitative comparison on four crack benchmarks.}
We report region metrics (mIoU/F1/Precision/Recall), boundary metric (ODS), and the connectivity-aware clDice~\cite{cldice}. Best results are highlighted in \textbf{bold}, and second-best are \underline{underlined}.}

\label{tab:sota_all}
\resizebox{\textwidth}{!}{%

\begin{tabular}{l cccccc cccccc}
\toprule

\multirow{2}{*}{Method} & \multicolumn{6}{c}{DeepCrack (\%)} & \multicolumn{6}{c}{Crack500 (\%)} \\

\cmidrule(lr){2-7} \cmidrule(lr){8-13}
 & mIoU & F1 & Prec & Recall & ODS & clDice & mIoU & F1 & Prec & Recall & ODS & clDice \\
\midrule
RIND~\cite{rind} & 81.27 & 84.40 & 82.22 & 86.64 & 90.72 & 95.69 & 78.24 & 72.98 & 70.85 & 79.41 & 72.98 & 80.93 \\
CT-CrackSeg~\cite{ctcrack} & 84.25 & 88.53 & 90.55 & 88.77 & 91.63 & \underline{96.46} & 77.81 & 74.06 & 75.65 & 72.53 & 74.38 & \underline{82.98} \\
SimCrack~\cite{simcrack} & 83.50 & 90.79 & 87.59 & 90.22 & 91.55 & 96.28 & \underline{78.76} & 74.35 & 74.55 & 78.25 & 74.42 & 82.03 \\
Swin-UMamba~\cite{swinumamba} & 83.52 & 83.72 & \underline{91.02} & 90.12 & \underline{91.74} & 95.91 & 77.52 & 72.86 & 74.05 & 76.37 & 73.02 & 80.53 \\
VMamba~\cite{vmamba_neurips} & 73.97 & 80.65 & 75.06 & 89.67 & 80.67 & 92.18 & 77.35 & 72.66 & 73.38 & 76.39 & 72.87 & 79.83 \\
VM-UNet~\cite{vmunet} & 80.19 & 88.51 & 87.74 & 90.83 & 88.54 & 94.36 & 76.83 & 71.58 & 71.05 & 77.22 & 71.60 & 79.16 \\
PlainMamba~\cite{plainmamba} & 73.47 & 82.65 & 88.26 & 89.17 & 82.89 & 91.14 & 70.94 & 62.74 & 66.74 & 65.34 & 63.36 & 66.48 \\
SCSegamba~\cite{scsegamba} & \underline{90.71} & \underline{91.20} & 90.39 & 91.00 & 91.70 & 94.48 & 78.75 & \underline{77.82} & \underline{76.27} & \underline{79.44} & \underline{77.82} & 79.50 \\
FFM~\cite{ffm} & 88.05 & 85.87 & 81.22 & \textbf{95.93} & 89.84 & 86.61 & 78.75 & 74.50 & 74.30 & 78.61 & 74.54 & 82.03 \\
GLCP~\cite{glcp} & 89.83 & 91.07 & 89.27 & 89.15 & 91.52 & 96.36 & 77.28 & 74.50 & 72.46 & 76.94 & 74.51 & 82.66 \\
\rowcolor[rgb]{.9,.9,.95}\textbf{\modelname{} (Ours)} & \textbf{91.29} & \textbf{91.43} & \textbf{91.36} & \underline{91.51} & \textbf{92.46} & \textbf{97.12} & \textbf{79.15} & \textbf{78.29} & \textbf{78.34} & \textbf{79.87} & \textbf{78.32} & \textbf{83.85} \\

\midrule
\midrule
\multirow{2}{*}{Method} & \multicolumn{6}{c}{CrackMap (\%)} & \multicolumn{6}{c}{TUT (\%)} \\
\cmidrule(lr){2-7} \cmidrule(lr){8-13}
 & mIoU & F1 & Prec & Recall & ODS & clDice & mIoU & F1 & Prec & Recall & ODS & clDice \\
\midrule
RIND~\cite{rind} & 76.44 & 65.25 & 58.95 & 86.42 & 75.04 & 80.52 & 80.52 & 80.50 & 76.51 & 86.48 & 80.74 & 91.08 \\
CT-CrackSeg~\cite{ctcrack} & \underline{79.93} & 73.36 & 60.09 & 85.74 & \underline{79.21} & 90.80 & 81.53 & 82.07 & 80.61 & 84.91 & 82.07 & \underline{91.31} \\
SimCrack~\cite{simcrack} & 79.78 & \underline{75.11} & 62.80 & 85.15 & 77.89 & \underline{91.80} & 81.90 & 81.61 & 78.31 & \underline{86.79} & 81.69 & 90.66 \\
Swin-UMamba~\cite{swinumamba} & 77.59 & 73.53 & 62.47 & 81.01 & 76.44 & 86.65 & 80.32 & 78.88 & 77.93 & 81.82 & 78.89 & 87.57 \\
VMamba~\cite{vmamba_neurips} & 77.96 & 65.07 & 50.70 & 82.64 & 67.16 & 69.28 & 79.45 & 77.82 & 75.72 & 82.11 & 77.85 & 87.68 \\
VM-UNet~\cite{vmunet} & 75.66 & 66.41 & 55.41 & 86.96 & 68.10 & 74.60 & 78.14 & 76.53 & 74.76 & 80.35 & 76.54 & 86.24 \\
PlainMamba~\cite{plainmamba} & 75.08 & 66.14 & 61.97 & 74.15 & 66.19 & 74.60 & 68.91 & 59.56 & 57.76 & 67.62 & 59.58 & 66.47 \\
SCSegamba~\cite{scsegamba} & 78.51 & 75.01 & 75.85 & 76.53 & 75.39 & 59.87 & 83.60 & \underline{82.21} & \underline{81.75} & 82.68 & \underline{82.21} & 84.99 \\
FFM~\cite{ffm} & 76.07 & 70.12 & 57.21 & \textbf{93.77} & 75.18 & 89.87 & \underline{84.62} & 81.88 & 80.43 & 84.70 & 81.88 & 91.09 \\
GLCP~\cite{glcp} & 70.30 & 69.80 & \textbf{76.33} & 78.35 & 70.34 & 68.37 & 80.74 & 76.12 & 69.46 & 85.41 & 77.34 & 87.52 \\
\rowcolor[rgb]{.9,.9,.95}\textbf{\modelname{} (Ours)} & \textbf{80.75} & \textbf{77.82} & \underline{76.27} & \underline{87.94} & \textbf{79.82} & \textbf{92.48} & \textbf{85.73} & \textbf{83.95} & \textbf{82.44} & \textbf{86.99} & \textbf{82.51} & \textbf{91.98} \\

\bottomrule
\end{tabular}%
}
\end{table*}

\subsection{Comparison with State-of-the-Art}
\label{subsec:sota}

\modelname{} consistently outperforms recent CNN, SSM, and topology-specialized baselines across all benchmarks (Tab.~\ref{tab:sota_all}).

\noindent\textbf{Mitigating Connectivity Breaks (DeepCrack \& Crack500).}
\modelname{} ranks 1st in ODS and clDice on both datasets, directly measuring boundary and topological integrity.
Notably, GLCP~\cite{glcp}---a recent topology-aware method that explicitly targets connectivity preservation via joint skeleton and local-discontinuity learning---achieves strong clDice on DeepCrack (96.36\%) but drops to 68.37\% on CrackMap, indicating that its supervisory strategy is sensitive to background clutter. In contrast, \modelname{} consistently achieves the highest clDice across all datasets, confirming that geometry-aligned serialization yields robust topological preservation.
FFM~\cite{ffm} attains the highest Recall on DeepCrack (95.93\%) but at the cost of substantially lower Precision (81.22\%) and clDice (86.61\%), revealing over-sensitivity to weak gradients---a typical symptom of texture leakage.

\noindent\textbf{Suppressing Texture Leakage (CrackMap \& TUT).}
On CrackMap, many methods exhibit ``high Recall, low Precision'' (Precision $\approx 50$--$57\%$), a signature of texture leakage. FFM exemplifies this pattern with 93.77\% Recall but only 57.21\% Precision.
\modelname{} achieves the highest mIoU (80.75\%) and clDice (92.48\%) while maintaining balanced Precision--Recall.
On TUT, spanning eight materially diverse environments, \modelname{} achieves the highest mIoU (85.73\%) and clDice (91.98\%), surpassing FFM (84.62\%) and GLCP (80.74\%), confirming that frequency-geometric disentanglement generalizes across diverse texture statistics.

\noindent\textbf{Summary.} FA-Scan alleviates connectivity breaks in low-contrast topology; ASGP suppresses texture leakage in cluttered environments. The combination outperforms not only general SSM backbones but also methods specifically designed for topological preservation~\cite{glcp} or self-similar structure encoding~\cite{ffm}.

\subsection{Model Efficiency}
\label{subsec:efficiency}

\begin{table*}[t]
\centering
\setlength{\tabcolsep}{8pt}
\renewcommand{\arraystretch}{0.7}
\footnotesize
\caption{\textbf{Efficiency analysis.}
All metrics are measured on a single NVIDIA RTX 3090 with input size $256{\times}256$ and batch size 1, averaged over 100 runs.
\modelname{} achieves a competitive accuracy–efficiency profile.}
\label{tab:eff}
\resizebox{0.75\linewidth}{!}{
\begin{tabular}{lccccc}
\toprule
Method & Year & Params$\downarrow$ & FLOPs$\downarrow$ & Size$\downarrow$ & FPS$\uparrow$ \\
\midrule
RIND~\cite{rind} & 2021 & 30.63M & 247.6G  & 117.06MB & 28.46 \\
CT-CrackSeg~\cite{ctcrack} & 2023 & 22.88M & 78.94G & 87.37MB & 66.0 \\
SimCrack~\cite{simcrack} & 2023 & 29.58M & 286.62G & 225MB & 40.81 \\
Swin-UMamba~\cite{swinumamba} & 2024 & 63.38M & 104.1G & 241.77MB & 43.1 \\
VMamba~\cite{vmamba_neurips} & 2024 & 19.50M & 395.24G & 65.54MB & 28.8 \\
VM-UNet~\cite{vmunet} & 2024 & 27.25M & 29.42G & 105.92MB & 27.9 \\
PlainMamba~\cite{plainmamba} & 2024 & 16.72M & 73.36G & 96MB & 90.6 \\
FFM~\cite{ffm} & 2024 & 55.43M & 260.07G & 211MB & 83.5 \\
SCSegamba~\cite{scsegamba} & 2025 & \textbf{3.05M} & 18.16G & 37MB & 17.9 \\
GLCP~\cite{glcp} & 2025 & 46.30M & 33.72G & 177MB & \textbf{202.6} \\
\rowcolor[rgb]{.9,.9,.95}\textbf{\modelname{} (Ours)} & - & 6.26M & \textbf{7.87G} & \textbf{23.92MB} & 80.2 \\
\bottomrule
\end{tabular}%
}
\end{table*}

% TABLE 3: SYSTEM DIAGNOSTICS (Updated)
% =================================================
\begin{table*}[t]
\centering
\caption{\textbf{System Diagnostics on DeepCrack.}
We conduct controlled ablations to verify:
(A) The impact of stage-adaptive gating policies in LGB;
(B) The sensitivity of ASGP to evolution steps ($T$) and probe count ($N$);
(C) The geometric sensitivity of different scan trajectories for specific frequency bands; and
(D) The stepwise contribution of each module, explicitly isolating the gains from the FA-Scan encoder versus the GFA/BRM decoder.}
\label{tab:diagnostics}

% --- Left Column: Table A ---
\begin{minipage}[t]{0.42\textwidth}
\vspace{0pt}
\centering
\renewcommand{\arraystretch}{1.485}
\setlength{\tabcolsep}{3pt}
    \resizebox{\linewidth}{!}{%
    \begin{tabular}{lccccc}
    \toprule
    \multicolumn{6}{c}{\textbf{A. LGB Gating Policy}} \\
    \midrule
    Policy & mIoU & F1 & P & R & ODS  \\
    \midrule
    None & 87.81 & 86.77 & 86.20 & 91.69 & 87.08 \\
    EEEE & 88.54 & 87.62 & 86.75 & 92.06 & 88.42  \\
    GGGG & 88.76 & 87.89 & 87.16 & \textbf{92.18} & 88.80 \\
    GEGE & 88.84 & 87.99 & 87.67 & 91.80 & 88.85  \\
    GGEE & 88.86 & 88.00 & 88.58 & 91.91 & 88.93  \\
    EGEG & 88.98 & 88.15 & 89.06 & 91.67 & 88.93  \\
    EGGE & 90.05 & 90.22 & 90.72 & 91.05 & 90.73  \\
    GEEG & 91.10 & 91.29 & 90.83 & 92.17 & 91.08  \\
    \rowcolor[rgb]{.9,.9,.95}\textbf{EEGG} & \textbf{91.29} & \textbf{91.43} & \textbf{91.36} & 91.51 & \textbf{92.46} \\
    \bottomrule
    \end{tabular}}
\end{minipage}
\hfill
% --- Right Column: Table B + C ---
\begin{minipage}[t]{0.56\textwidth}
    \vspace{0pt}
    \centering
    \renewcommand{\arraystretch}{0.9}

    % --- Table B: ASGP Hyper-parameters ---
\resizebox{\linewidth}{!}{%
\begin{tabular}{lccccccc}
\toprule
\multicolumn{7}{c}{\textbf{B. ASGP Hyper-parameters}} \\
\midrule
Setting & mIoU & F1 & P & R & ODS & FPS$\uparrow$ \\ 
\midrule
\rowcolor[rgb]{.9,.9,.95}\textbf{$T{=}3,N{=}64$} & \textbf{91.29} & \textbf{91.43} & \textbf{91.36} & 91.51 & \textbf{92.46} & \textbf{80.2} \\ 
$T{=}3,N{=}32$ & 90.80 & 90.20 & 89.52 & 91.84 & 91.56 & 81.5 \\
$T{=}1,N{=}64$ & 88.93 & 88.77 & 85.93 & \textbf{92.91} & 89.26 & 82.1 \\
$T{=}7,N{=}64$ & 88.30 & 87.39 & 86.77 & 91.34 & 88.66 & 76.4 \\
$T{=}3,N{=}128$ & 87.78 & 86.55 & 85.82 & 90.21 & 86.72 & 79.8 \\
\bottomrule
\end{tabular}}

    \vspace{4pt}

    % --- Table C: Scan Trajectory (Split Columns) ---
    \setlength{\tabcolsep}{1.2pt}
    \resizebox{\linewidth}{!}{%
    \begin{tabular}{llcccccc}
    \toprule
    \multicolumn{8}{c}{\textbf{C. Scan Trajectory Ablation (FA-Scan)}} \\
    \midrule
    Row & \textbf{HH Path} & LH / HL Path & mIoU & F1 & P & R & ODS \\
    \midrule
    \rowcolor[rgb]{.9,.9,.95}D1 & \textbf{Hilbert} & LH$\to$H, HL$\to$V & \textbf{90.46} & \textbf{89.63} & \textbf{89.46} & \textbf{89.45} & \textbf{89.73} \\
    D2 & \textbf{Raster} & LH$\to$H, HL$\to$V & 89.12 & 88.34 & 88.22 & 88.06 & 88.51 \\
    D3 & Z-order & LH$\to$H, HL$\to$V & 90.05 & 89.18 & 89.02 & 88.90 & 89.10 \\
    D4 & Hilbert & LH→V, HL→H & 88.74 & 87.90 & 87.68 & 87.83 & 87.96 \\
    D5 & Hilbert & Snake (Bi-dir)     & 89.82 & 88.95 & 88.80 & 88.62 & 88.93 \\
    D6 & Hilbert & Hilbert (All)      & 89.46 & 88.60 & 88.42 & 88.33 & 88.66 \\
    \bottomrule
    \end{tabular}}
\end{minipage}

\vspace{2pt}

% --- Bottom Panel: Table D ---
\renewcommand{\arraystretch}{1.05}
\setlength{\tabcolsep}{4pt}
\resizebox{1.0\textwidth}{!}{%
\begin{tabular}{lccccccccc}
\toprule
\multicolumn{9}{c}{\textbf{D. Mechanism Dissection}} \\
\midrule
Row & Variant & mIoU & F1 & P & R & ODS & Params$\downarrow$ & FLOPs$\downarrow$ \\
\midrule
1 & Standard Mamba Encoder & 85.98 & 86.11 & 86.70 & 86.87 & 86.22 & 2.20M & 7.81G \\
\rowcolor{gray!15}2 & VMamba Encoder + GFA/BRM & 87.56 & 88.12 & 87.45 & 89.30 & 88.65 & 21.50M & 12.40G \\
3 & Baseline (DWT + Raster) & 86.44 & 86.97 & 87.09 & 87.26 & 86.97 & 6.09M & 7.80G \\
\rowcolor{gray!15}4 & DWT + Cross-Scan (4-dir) & 88.15 & 87.62 & 87.90 & 88.48 & 87.95 & 6.10M & 8.85G \\
5 & Static Gate ($M_0$ Only) & 88.67 & 87.12 & 88.29 & 89.13 & 88.08 & 6.26M & 7.87G \\
6 & ASGP ($T{=}3$, w/o FA-Scan) & 89.83 & 88.93 & 88.51 & 89.36 & 89.63 & 6.22M & 7.87G \\
7 & FA-Block (FA-Scan + LGB) & 90.46 & 89.63 & 89.46 & 89.45 & 89.73 & 6.09M & 7.80G \\
8 & \modelname{} (w/o Align) & 90.75 & 90.62 & 90.31 & 90.98 & 91.35 & 6.25M & 7.86G \\
\rowcolor[rgb]{.9,.9,.95}9 & \textbf{\modelname{} (Full)} & \textbf{91.29} & \textbf{91.43} & \textbf{91.36} & \textbf{91.51} & \textbf{92.46} & 6.26M & 7.87G \\
\bottomrule
\end{tabular}}
\end{table*}

% =========================================================
\modelname{} achieves a strong accuracy--efficiency profile (Tab.~\ref{tab:eff}).
Compared to Swin-UMamba, it reduces FLOPs by $>$90\% and doubles throughput (80.2 FPS). SCSegamba is more compact (3.05M) but only reaches 17.9 FPS; \modelname{} delivers 4.5$\times$ the speed at 56\% fewer FLOPs.
Notably, GLCP~\cite{glcp} achieves the highest raw FPS (202.6) through optimized operators, yet requires 7.4$\times$ more parameters and 4.3$\times$ more FLOPs than \modelname{} while lagging in accuracy by up to 10.5\% mIoU (CrackMap). FFM~\cite{ffm} demands 33$\times$ more FLOPs (260G).
Crucially, FA-Scan adds no parameters (index reordering only), and ASGP's sparse probing ($N{=}64$) incurs negligible overhead ($\sim$3 ms per stage).

\subsection{Ablation Studies and Diagnostics}
\label{subsec:ablation}

\begin{figure*}[t]
 \centering
 \includegraphics[width=0.80\linewidth]{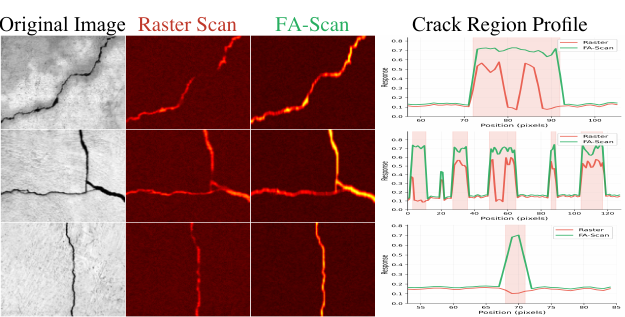}

 \caption{\textbf{Feature Response Analysis: Raster vs. FA-Scan.}
 Feature maps are extracted from the Stage-2 encoder output (before IDWT), and profiles are sampled along the crack structure (highlighted in pink).
 The centerline is derived from the ground-truth skeleton for visualization purposes.
 Raster Scan: Features exhibit sharp signal drops when the scan path cuts across the crack, leading to fragmentation.
 FA-Scan: By aligning the scan trajectory with sub-band orientation, our method maintains a continuous, high-amplitude response (measured as the
$\ell_2$-normalized mean channel activation), verifying the preservation of
connectivity.}
 \label{fig:scan_diag}
\end{figure*}

To validate our gains stem from frequency-geometric disentanglement and active probing rather than capacity scaling, we ablate key components on DeepCrack.

\noindent\textbf{Feature Response Analysis (\figref{fig:scan_diag}).}
\figref{fig:scan_diag} visualizes the 1D feature response along a crack: raster scanning (red) shows severe signal dropouts where the path cuts across the structure, while FA-Scan (green) maintains a continuous, high-amplitude trace.

\noindent\textbf{H1: Disentangling Architecture vs.\ Serialization (Tab.~\ref{tab:diagnostics}C, D).}
A central question is whether performance gains stem from the DWT multi-branch architecture or from geometry-aligned serialization. We answer this through a controlled causal chain with an explicit scan-diversity baseline.

\noindent\textbf{1) DWT decomposition alone is insufficient.}
Introducing DWT without scan alignment (Row~1$\to$Row~3) yields a negligible +0.46\% mIoU, confirming that the multi-branch frequency architecture is not the performance driver.

\noindent\textbf{2) Scan diversity helps, but alignment helps far more.}
Replacing single-direction Raster with VMamba-style Cross-Scan (Row~3$\to$Row~4), which applies four-directional raster to each DWT sub-band using the same shared SSM, improves mIoU by +1.71\% but increases FLOPs to 8.85G. In contrast, FA-Scan (Row~3$\to$Row~7) achieves a much larger +4.02\% gain at lower FLOPs (7.80G).
This indicates that frequency–geometry alignment contributes substantially more than directionality alone, while being computationally cheaper.

\noindent\textbf{3) Alignment, not diversity, is the core driver.}
Tab.~\ref{tab:diagnostics}C provides fine-grained confirmation:
deliberately mismatching sub-band assignments (D4) drops mIoU by $-1.72\%$ despite using the same number of scan directions;
uniformly applying Hilbert (D6, 89.46\%) or Snake scanning (D5, 89.82\%)---both multi-directional strategies---also underperform the aligned FA-Scan (D1, 90.46\%).
These results suggest that the gains mainly arise from aligning trajectories with sub-band geometry rather than from scan diversity itself.

\noindent\textbf{H2: Static vs. Active Probing (Tab.~\ref{tab:diagnostics}B, D).}
We decompose ASGP's contribution into two stages.
First, the transition from no gating (Row~3, 86.44\%) to static gating (Row~5, 88.67\%) contributes +2.23\% mIoU, confirming that topology-conditioned filtering of high-frequency noise is beneficial even without evolution.
Second, adding gradient-guided evolution (Row~5$\to$Row~6) further improves mIoU by +1.16\% and ODS by +1.55\%, demonstrating that iterative probe refinement meaningfully enhances gate quality.
However, excessive steps ($T{=}7$) over-smooth features, degrading all metrics (Tab.~\ref{tab:diagnostics}B).
Additionally, removing the diversity constraint ($\lambda_2$) causes probes to collapse into high-response centroids; the truncated repulsion force effectively disperses them along curvilinear ridges to ensure complete topological coverage.

\noindent\textbf{H3: Gating Policy (Tab.~\ref{tab:diagnostics}A).}
The hybrid EEGG policy balances preserving weak details via ECA in shallow layers (maximizing Recall) and rejecting false positives via GSE in deeper layers (maximizing Precision), yielding the optimal 91.29\% mIoU trade-off.

\noindent\textbf{H4: Encoder vs. Decoder Contribution (Tab.~\ref{tab:diagnostics}D).}
A strong baseline of VMamba Encoder + GFA/BRM (Row~2) lags behind the full \modelname{} (Row~9) by 3.73\% mIoU and 3.81\% ODS, confirming that encoder-side geometry alignment provides structural cues the decoder alone cannot recover.

\noindent\textbf{H5: Necessity of Spatial Alignment (Tab.~\ref{tab:diagnostics}D).}
Removing the detail-driven \texttt{GridSample} alignment (Row 8) forces the network to process uncalibrated features. This drops ODS by 1.11\% (92.46\%$\to$91.35\%) and Precision by 1.05\%. Without alignment, spatial shifts from downsampling cause high-frequency noise to misalign with the low-frequency skeleton. The Align module corrects this dislocation, providing a geometric foundation.

\subsection{Qualitative Results}
\label{subsec:qual}

\begin{figure*}[t]
  \centering
  \includegraphics[width=1\textwidth]{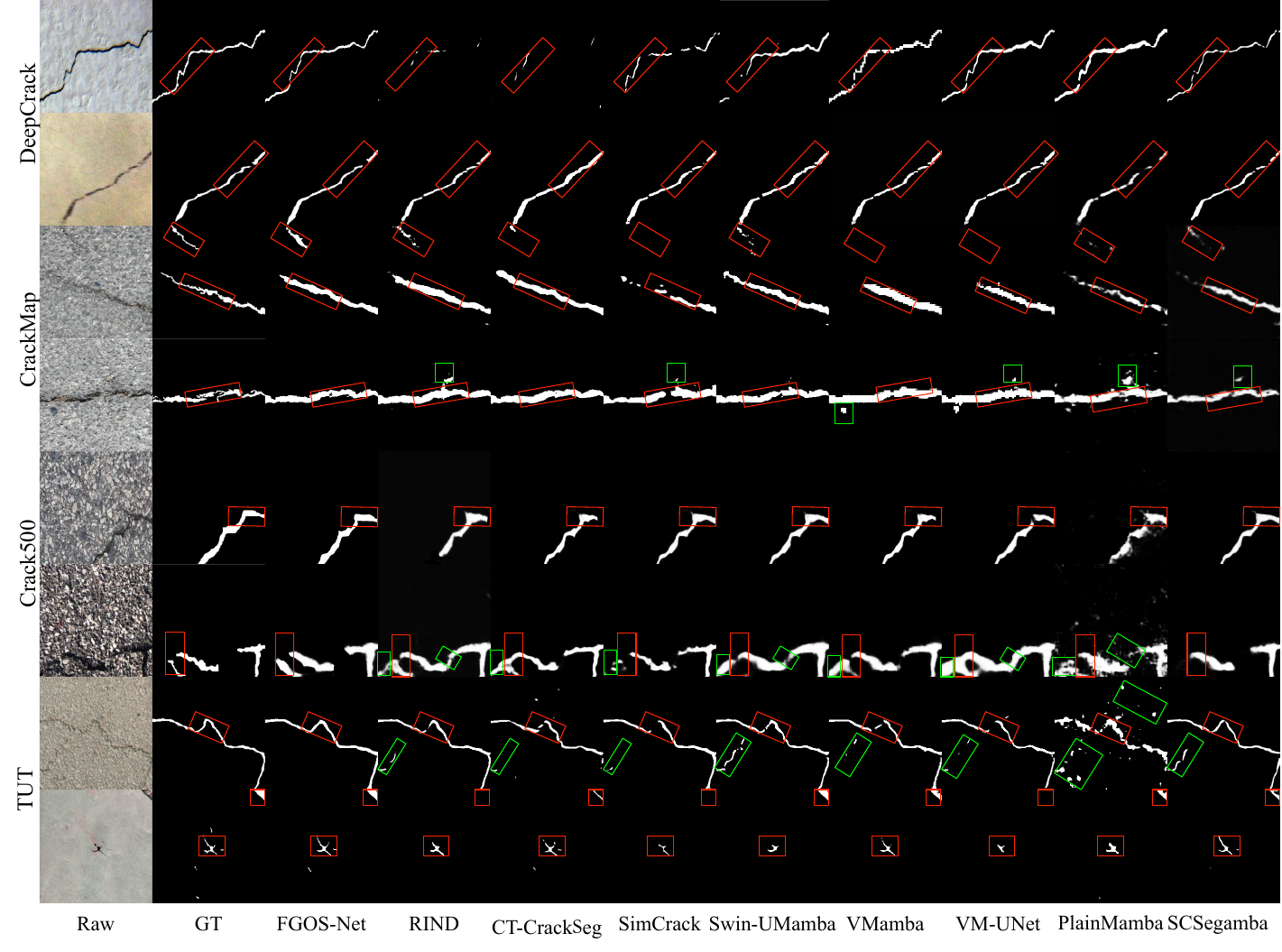}
  \caption{\textbf{Qualitative comparison on challenging scenarios.}
  We visualize predictions from representative CNN/SSM baselines against \modelname{} (Ours). All methods are visualized on the same test samples.
  Red Boxes (Topology): In low-contrast regions, baselines suffer from connectivity breaks, creating fragmented masks. Ours maintains continuous traces.
  Green Boxes (Texture): Under heavy clutter (\eg,  water stains, oil spots), baselines yield false positives (texture leakage). Ours successfully suppresses these artifacts via topology-gated injection.}
  \label{fig:qual}
\end{figure*}

\noindent\textbf{Mitigating Connectivity Breaks (Red Boxes).}
As shown in the top rows of \cref{fig:qual}, faint cracks often fade into the background, challenging the long-range modeling capability of standard backbones.
Isotropic serialization-based approaches fail to aggregate weak signals along the crack direction, resulting in fragmented, spotty predictions.
In contrast, \modelname{} reconstructs continuous crack paths even when local contrast is minimal.
This visual continuity directly confirms FA-Scan's structure-aligned propagation.

\noindent\textbf{Suppressing Texture Leakage (Green Boxes).}
In texture-heavy scenes (bottom rows), background artifacts such as water stains and pavement grain create ambiguous high-frequency gradients.
Baselines lacking explicit frequency disentanglement tend to over-activate on these regions, leading to severe false positives (texture leakage).
\modelname{} produces notably clean masks.
This confirms ASGP's gating: non-structural clutter is filtered while skeleton-aligned gradients are preserved.

\subsection{Cross-Domain Generalization.}

To verify that \modelname{} generalizes beyond crack detection,
we directly apply the same architecture—using identical
hyperparameters and $256{\times}256$ inputs—to two distinct
thin-structure domains: retinal vessel segmentation
(CHASEDB1~\cite{chase}, 28 images) and aerial road extraction
(Massachusetts Roads~\cite{mass_roads}, 1171 images).

As shown in Tab.~\ref{tab:cross}, \modelname{} consistently
outperforms representative CNN and SSM baselines from the
crack benchmarks without any domain-specific tuning.
In particular, the improvement in clDice is notable (+6.05\% on vessels and +7.46\% on roads relative
to SCSegamba), demonstrating that geometry-aligned
serialization and topology-conditioned gating effectively
preserve long-range continuity across diverse thin-structure modalities.
\begin{table}[t]
\centering
\footnotesize
\setlength{\tabcolsep}{2pt}
\renewcommand{\arraystretch}{1.05}

\caption{\textbf{Cross-domain generalization.}
All models follow identical training settings. Best results are highlighted in \textbf{bold}.}
\label{tab:cross}

\begin{tabular}{lcccccccc}
\toprule
\multirow{2}{*}{Method}
& \multicolumn{4}{c}{CHASEDB1 (Vessels) [\%]}
& \multicolumn{4}{c}{Massachusetts (Roads) [\%]} \\
\cmidrule(lr){2-5} \cmidrule(lr){6-9}
& mIoU & F1 & Prec & clDice
& mIoU & F1 & Prec & clDice \\
\midrule

SimCrack~\cite{simcrack}
& 75.82 & 74.11 & 69.37 & 75.24
& 72.63 & 65.81 & 75.10 & 71.94 \\

VMamba~\cite{vmamba_neurips}
& 73.00 & 70.32 & 57.21 & 76.01
& 71.51 & 62.89 & 73.78 & 71.20 \\

SCSegamba~\cite{scsegamba}
& 78.17 & 76.65 & 67.25 & 77.30
& 75.41 & 64.22 & 76.14 & 73.40 \\

\rowcolor[rgb]{.9,.9,.95}
\textbf{\modelname{} (Ours)}
& \textbf{80.45} & \textbf{79.12} & \textbf{76.71} & \textbf{83.35}
& \textbf{79.83} & \textbf{69.32} & \textbf{80.99} & \textbf{80.86} \\

\bottomrule
\end{tabular}
\end{table}

\section{Conclusion}
\label{sec:conclusion}

We identified a key bottleneck in applying State-Space Models to thin-structure segmentation: the geometry mismatch introduced by isotropic serialization, which disrupts connectivity and introduces texture leakage.
To address this issue, we proposed \modelname{}, a framework based on frequency--geometric disentanglement that separates coarse topology from directional details via DWT and couples them through geometry-aligned FA-Scan and topology-gated ASGP.
Extensive experiments on four crack benchmarks and two cross-domain datasets show that \modelname{} achieves strong boundary integrity while maintaining a competitive accuracy--efficiency trade-off.

Although evaluated on crack segmentation as an extreme testbed, the proposed frequency-geometric disentanglement---aligning serialization with sub-band geometry and conditioning detail injection on topology---is architecture-agnostic and extensible to other thin-structure domains (\eg, retinal vessels and road networks) where similar serialization mismatches arise.

\section{Acknowledgments}
This work was supported by the National Key Research and Development Program of China under Grant 2023YFB3906102.

\newpage
\bibliographystyle{splncs04}
\bibliography{main}

\end{document}